\documentclass[runningheads]{llncs}
\usepackage[mobile]{eccv}
\usepackage{colortbl}
\usepackage{eccvabbrv}
\usepackage{graphicx}
\usepackage{booktabs}
\usepackage{multirow}
\usepackage{tabularx}
\usepackage{pifont}
\usepackage{amsmath}
\usepackage[T1]{fontenc}
\usepackage[sectionbib]{chapterbib}
\usepackage{arydshln}
\usepackage{wrapfig}
\usepackage{microtype}
\usepackage[accsupp]{axessibility} 
\usepackage{hyperref}

\definecolor{cvprblue}{rgb}{0.21,0.49,0.74}
\definecolor{lightgray}{rgb}{0.835, 0.835, 0.835}
\definecolor{lightergray}{rgb}{0.935, 0.935, 0.935}
\definecolor{lighterblue}{rgb}{0.93, 0.97, 1.0}
\definecolor{lightblue}{rgb}{0.83, 0.90, 1.0}

\newcommand{\cmark}{\ding{51}}
\newcommand{\xmark}{\ding{55}}
\newcommand\aug{\fboxsep=-\fboxrule\fbox{\strut}}

\DeclareMathOperator*{\argmin}{arg\,min}
\DeclareMathOperator{\diag}{diag}
\begin{document}

\title{HandDGP: Camera-Space Hand Mesh Prediction \\ with Differentiable Global Positioning} 
\author{Eugene Valassakis\thanks{Now at Synthesia. Work done while at Niantic.}, Guillermo Garcia-Hernando}

\institute{Niantic\\
\url{https://nianticlabs.github.io/handdgp/}}

\authorrunning{Eugene Valassakis, Guillermo Garcia-Hernando}

\titlerunning{HandDGP}

\maketitle

\begin{abstract}
Predicting camera-space hand meshes from single RGB images is crucial for enabling realistic hand interactions in 3D virtual and augmented worlds. Previous work typically divided the task into two stages: given a cropped image of the hand, predict meshes in relative coordinates, followed by lifting these predictions into camera space in a separate and independent stage, often resulting in the loss of valuable contextual and scale information. To prevent the loss of these cues, we propose unifying these two stages into an end-to-end solution that addresses the 2D-3D correspondence problem. This solution enables back-propagation from camera space outputs to the rest of the network through a new differentiable global positioning module. We also introduce an image rectification step that harmonizes both the training dataset and the input image as if they were acquired with the same camera, helping to alleviate the inherent scale-depth ambiguity of the problem. We validate the effectiveness of our framework in evaluations against several baselines and state-of-the-art approaches across three public benchmarks. 
\keywords{camera-space hand mesh estimation \and hand and body pose shape from RGB images \and 3D-to-2D scale ambiguity \and differentiable solver}
\end{abstract}    

\section{Introduction}
\label{sec:intro}

Predicting 3D hand meshes from single-view RGB images has become an increasingly popular research area due to its potential in augmented and virtual reality applications, such as virtual try-on experiences~\cite{tang2021towards}, human digitization~\cite{chen2023hand},  gaming~\cite{han2020megatrack}, and teleoperation~\cite{antotsiou2018task, garcia2020physics}. Despite recent progress, challenges remain~\cite{armagan2020measuring, yuan2018depth} due to the hand's articulated structure, self-occlusions, annotation difficulty, and 2D-to-3D scale and depth ambiguity.

Because of these challenges, most previous work have focused on predicting quality root-relative hand meshes, \ie 3D hand meshes in coordinates relative to a pre-defined root joint, such as the wrist~\cite{chen2022mobrecon}, as opposed to predicting in the global camera space. Root-relative predictions with a camera projection model~\cite{kanazawa2018end, zhang2019end, boukhayma20193d, baek2019pushing} can be sufficient in applications that end up displayed on 2D images, such as virtual try-on experiences. However, camera-space predictions are critical for interactions in 3D virtual and augmented worlds, \eg in applications such as gaming, and office work when using mixed-reality headsets~\cite{quest3, AVP}.

\begin{figure}[t]
    \centering
    \includegraphics[width=\textwidth]{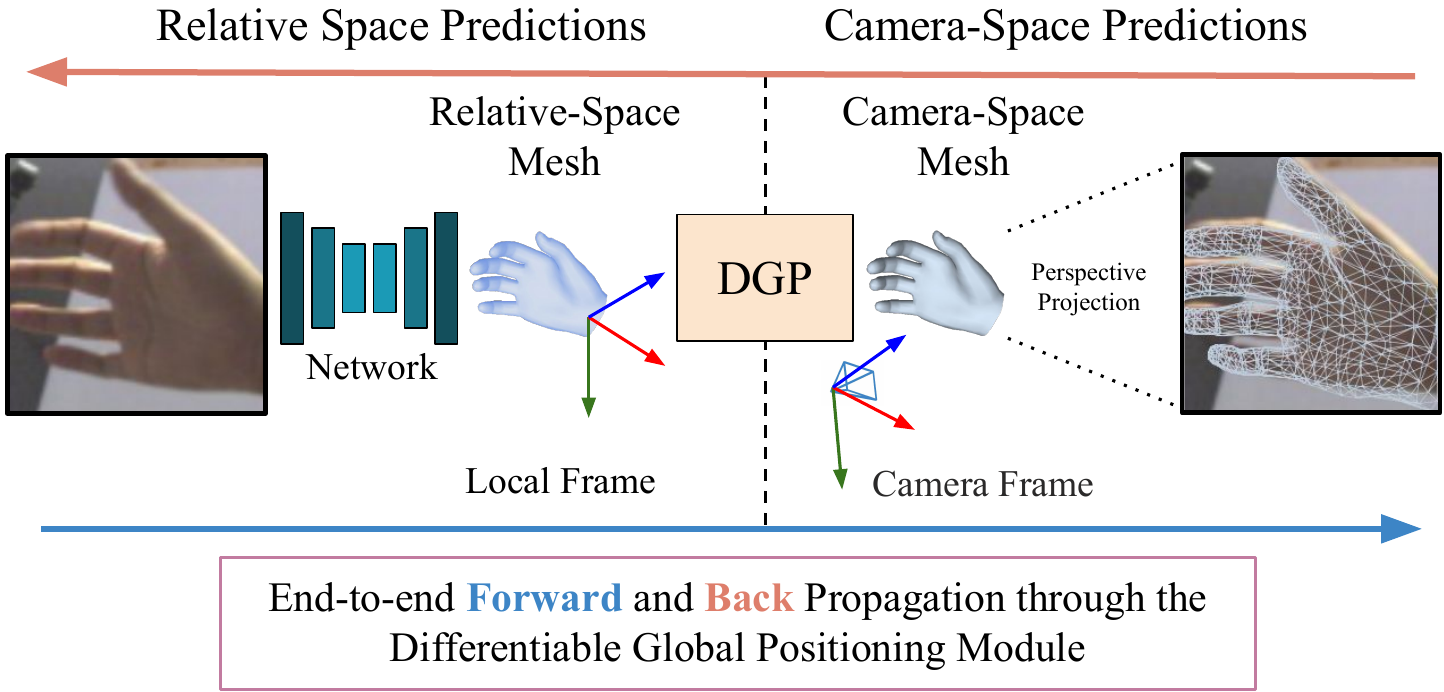}
    \caption{\textbf{Method overview:} Through our Differentiable Global Positioning module (DGP) which predicts the root translation of the hand, our method is able to back-propagate through the root-finding operation, enabling an end-to-end solution.}
    \label{fig:showcase}
\end{figure}

For tasks that need 3D camera-space hand meshes, the dominant approach is to take those root-relative hand meshes and then lift those predictions to the camera-space 3D coordinates in a separate, independent process. For example, Iqbal~\etal~\cite{iqbal2018hand} and Zhang~\etal~\cite{zhou2020monocular} output predictions in a relative 2.5D space and infer the global coordinates analytically up to scale. With a similar relative representation, Moon~\etal~\cite{I2L} and Tang~\etal~\cite{tang2021towards} predict camera-space coordinates with the aid of an independent network known as RootNet~\cite{RootNet}. Exploiting 2D-to-3D correspondences, CMR~\cite{chen2021camera} and MobRecon~\cite{chen2022mobrecon} first predict both 2D keypoints and 3D root-relative meshes, and then find the 3D rigid transformation that best explains the mesh projection via an independent test-time optimization process. Similarly, HandOccNet~\cite{park2022handoccnet} also uses a test-time optimization to predict scale and root translation by minimizing a 2D projection loss. All these methods confine the learning stage to the relative space, yielding state-of-the-art relative meshes with high efficiency, but falling short when it comes to placing the hand in the camera space, as shown in our experiments.

We propose to achieve the best of both worlds by simultaneously learning root-relative meshes and the 3D lifting function in an end-to-end manner. To this end, we propose a Differentiable Global Positioning (DGP) module, a modern take on the classical Direct Linear Transform algorithm~\cite{hartley2003multiple}. DGP enables the backpropagation of gradients directly from camera space outputs to the 2D-3D correspondences defined by the 2D keypoint predictions and the root-relative 3D hand meshes. 
Thanks to being differentiable, DGP could potentially also be included in the learning of any hand mesh prediction neural network that predicts 2D hand keypoints and root-relative 3D hand meshes such as CMR~\cite{chen2021camera} and MobRecon~\cite{chen2022mobrecon}. In our experiments, we show that allowing gradients to flow from the camera-space through the root-relative network results in better global hand mesh predictions compared to disjoint two-step approaches, \eg using RootNet~\cite{RootNet} or test-time optimizations, as well as an end-to-end regression baseline that predicts both camera-space translation and relative predictions. Further, inspired by recent metric 3D scene geometry prediction work~\cite{yin2023metric3d}, we found in our experiments that a simple image and camera parameter rectification step helps to alleviate the inherent 2D-to-3D depth and scale ambiguity problem. The core idea is to conduct all learning as though the images were captured using the same camera model, thereby reducing the ambiguity that the network must resolve. While this approach leads to improved camera-space predictions, it incurs a slight decline in performance for relative-space predictions. We thoroughly examine the impact of this rectification step on both our method and on baseline approaches. 
In summary, our \textbf{contributions} are as follows: 
\begin{itemize}
    \item We propose HandDGP, a flexible framework that unifies the learning of both root-relative and camera-space hand meshes in an end-to-end manner.
    \item We identify and highlight the importance of performing image rectification in alleviating some of the 2D-to-3D depth and scale ambiguity in the camera-space hand mesh prediction problem.
    \item We conduct an extensive experimental evaluation of various design choices for addressing the problem of camera-space hand mesh prediction, aiming to encourage research in this direction. This area has been somewhat overlooked in previous work in favor of root-relative predictions.
\end{itemize}
\section{Related Work}
\label{sec:related_work}
\paragraph{Camera-Space 3D Hand Mesh Prediction.} Most previous works in monocular RGB-based camera-space 3D hand mesh and pose estimation follow a two-stage approach: (1) a first stage for predicting hand mesh/pose in root-relative coordinates, and (2) a lifting stage to recover camera-space coordinates from root-relative ones. For monocular 3D hand pose estimation, Iqbal \etal~\cite{iqbal2018hand} predicts in 2.5D root-relative space, and then lifts those predictions to 3D camera space predictions using an analytical solution. This result is up to a scaling factor, however, and to resolve the scale ambiguity an extra scale parameter is required, which is assumed to be provided~\cite{spurr2020weakly} or globally estimated from data. I2L-MeshNet~\cite{I2L} proposes a regression approach to recover root-relative 2.5D meshes and subsequently lifts them to camera-space using a separate network  RootNet~\cite{RootNet} which uses prior anthropometric knowledge to reduce scale ambiguity~\cite{li2022cliff}. NFV~\cite{huang2023neural} proposes a neural voting approach with a 3D implicit function that directly regresses 3D hand poses in camera-space from full images. NFV uses a Marching Cubes post processing to predict meshes, which degrades efficiency and is at the cost of not having semantic mappings for their predicted vertices, which are crucial for some applications. Hasson~\etal~\cite{hasson2020leveraging} predicts both object and hand camera-space translations using both hand and object cues, but makes assumptions on the geometry of the object which facilitate the scale recovery. Closest to our work, CMR~\cite{chen2021camera},  MobRecon~\cite{chen2022mobrecon} and HandOccNet~\cite{park2022handoccnet} first predict both 2D keypoints and 3D root-relative meshes, while 3D camera-space coordinates are obtained with a test-time registration function that estimates the root position. This function typically aims to find the 3D rigid transformation for the hand mesh that best projects into the 2D input image given correspondences between 2D hand keypoints and 3D mesh keypoints. Our method builds on top of such 2D-3D paradigm with the key difference of leveraging a differentiable registration function, enabling us to directly learn our mesh network directly in the camera-space in an end-to-end manner. We compare our work with \cite{I2L, RootNet, huang2023neural, hasson2020leveraging, chen2021camera, chen2022mobrecon, park2022handoccnet} in Section~\ref{sec:sota} and find that our method outperforms these state-of-the-art methods in camera-space predictions.

\paragraph{Root-Relative 3D Hand Mesh Prediction.} Different methods have been proposed for RGB-based monocular hand mesh reconstruction, with various hand output representations, such as parametric models, voxels, vertices, implicit functions or UV maps~\cite{chen2021i2uv}. Parametric models~\cite{hasson19_obman, boukhayma20193d,baek2019pushing,baek2020weakly, zhang2021hand, chen2021camera, chen2021model, cao2021reconstructing, hampali2022keypoint} typically regress hand shape and pose coefficients of the MANO parametric hand model~\cite{romero2022embodied}. These methods are intrinsically limited by the expressiveness of the model used. Model-free methods circumvent these limitations by working on high dimensional representations. This can be true in voxel-based methods, such as I2L-MeshNet~\cite{I2L}, which represents volumetric hand data in a 2.5D manner at the cost of efficiency. Also typically constrained by efficiency but at the benefit of high resolution are implicit function methods~\cite{karunratanakul2020grasping,karunratanakul2021skeleton,huang2023neural}, which inherit from the trend started by human body digitization~\cite{saito2019pifu, mihajlovic2021leap, bhatnagar2020loopreg, peng2021neural, huang2023neural}. Vertex-based methods~\cite{ge20193d,kulon2020weakly,lin2021end, chen2021camera, lin2021mesh, chen2022mobrecon, park2022handoccnet} aim to directly predict 3D vertex coordinates. In this work we build upon MobRecon's framework~\cite{chen2022mobrecon} by adopting their pipeline: predicting 2D keypoints following an encoder-decoder approach then leveraging those keypoints to grid-sample features that are lifted to root-relative 3D by a graph neural decoder. We extend this framework by instead learning in camera-space owing to our proposed differentiable global positioning function. In Section~\ref{sec:baselines} we show that our method improves \cite{chen2022mobrecon}'s camera-space predictions significantly.

\paragraph{Differentiable Correspondence Solvers.} Solving for unknown geometric quantities using 2D-2D or 2D-3D  correspondences has long been a central subject in computer vision~\cite{prince2012computer,hartley2003multiple}. With the emergence of deep learning, several studies have attempted to integrate these well known geometric and algebraic solutions to deep learning pipelines~\cite{remelli2020lightweight,chen2020end, brachmann2017dsac, bhowmik2020reinforced, wei2023generalized,epropnp}. Notable examples include (1) Chen \etal.~\cite{chen2020end} which propose a differentiable perspective-n-point (PnP) solver and validate it in various problems such as pose estimation, or camera calibration, and Remelli \etal~\cite{remelli2020lightweight} that use a differentiable Direct Linear Transform (DLT) implementation to perform multi-view body pose estimation. We derive a DLT solution to the root finding problem in the context of hand-mesh inference, and show that it can be implemented differentiably and integrated to an end-to-end pipeline for camera-space hand-mesh prediction.
\clearpage
\begin{figure*}[t]
    \centering
    \includegraphics[width=1\textwidth]{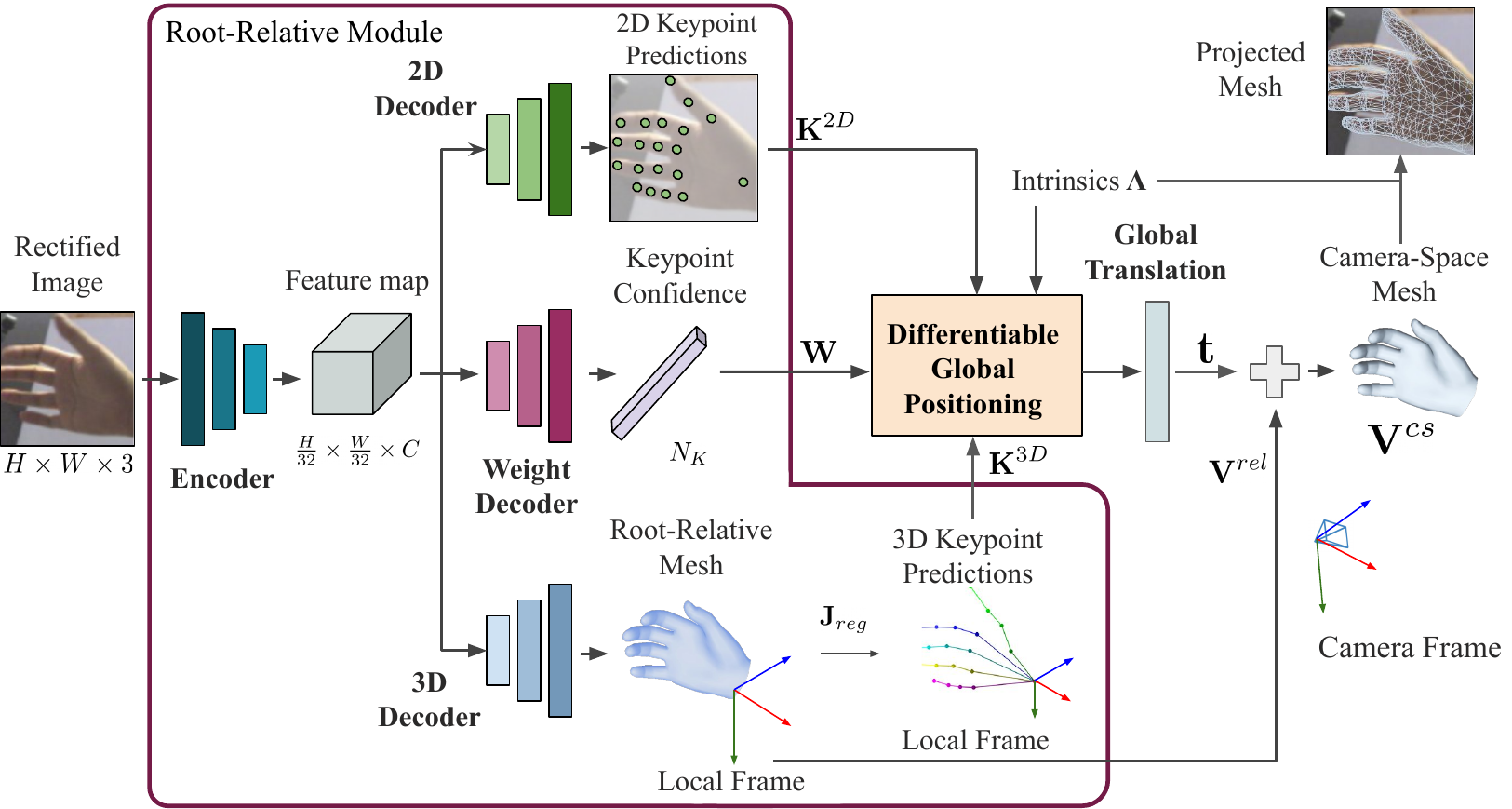}
    \caption{\textbf{HandDGP Framework Overview.} Rectified images are passed through our framework, which predicts camera-space coordinates using our proposed DGP module.}
    \label{fig:main_fig} 
\end{figure*}
\
\section{Proposed Framework}
\label{sec:method:overview}
\noindent\textbf{Overview.} Shown in Figure~\ref{fig:main_fig}, the core idea behind our approach is to exploit the geometry of the problem to integrate hand root finding in a  differentiable pipeline that can predict a hand mesh directly in 3D camera-space coordinates.\\

\noindent\textbf{Predicting root-relative hand meshes.} Starting from one RGB image $\mathbf{I}\in\mathbb{R}^{H \times W \times 3}$, we first predict a set of 2D keypoints $\mathcal{\mathbf{K}}^{2D}=\{k^{2D}_i\}_{i=1}^{N_K}$, that can be joints or other landmarks, a set of root-relative 3D vertices $\mathbf{V}^{rel}=\{v_i^{rel}\}_{i=1}^{N_V}$, and a set of weights $\mathbf{W}=\{w_i\}_{i=1}^{N_K}$, that represent the confidence in the predictions of each keypoint. We then obtain $\mathbf{K}^{3D}=\{k_i^{3D}\}_{i=1}^{N_K}$, a set of root-relative $3D$ keypoints on the hand model that correspond to the 2D keypoints $\mathbf{K}^{2D}$. To obtain $\mathbf{K}^{3D}$, we assume having access to a 3D keypoint regressor $\mathbf{J}_{reg}: \mathbf{V}^{rel}\rightarrow \mathbf{K}^{3D}$.  $\mathbf{J}_{reg}$ usually comes in the form of a matrix, which defines keypoints on the hand as a linear combination of mesh vertices, and is typically provided with popular mesh models such as MANO~\cite{romero2022embodied} for hands and SMPL~\cite{pavlakos2019expressive} for full body meshes.\\ 
\noindent\textbf{Finding the hand root with HandDGP.} Our key innovation is our differentiable global positioning (DGP or positioning module, for brevity), which uses the root-relative 3D keypoints  $\mathbf{K}^{3D}$, the 2D keypoints $\mathbf{K}^{2D}$ and the weights $\mathbf{W}$ in order to predict a global translation $\mathbf{t}\in\mathbb{R}^3$ in camera-space which we can use to obtain the camera-space vertex predictions $\mathbf{V}^{cs} = \{v_i^{cs}\}_{i=1}^{N_V}$, with
\begin{equation}\label{sec:method:eq_vcs}
    v_i^{cs}= v_i^{rel}+\mathbf{t}.
\end{equation}

The camera-space vertex predictions can finally be used to project the mesh into 2D using a pinhole camera perspective projection. This pipeline allows us to include the global root translation and the resulting mesh projections as part of our network training. Incorporating  root prediction as part of the training this way has the benefit of avoiding the accumulation of errors that can occur when using two independent processes for root-relative predictions and root-finding.  We finally note that (1) our method is agnostic to the particular method used to obtain the predictions for $\mathbf{V}^{rel}$, $\mathbf{K}$ and $\mathbf{W}$ and (2) the task of recovering the root of the hand is equivalent to finding the hand's global translation in camera space. Therefore, we will use both terms interchangeably throughout.

\subsection{Differentiable Global Positioning}
\label{sec:method:DLT}
At the core of our method lies the differentiable global positioning module, which takes in  $\mathbf{K}^{3D}$,  $\mathbf{K}^{2D}$, $\mathbf{J}_{reg}$, and the camera intrinsics $\mathbf{\Lambda}$ as input, and  outputs the global camera-space translation $\mathbf{t}$ in a differentiable manner. Although our full approach also considers the keypoint confidences $\mathbf{W}$, for clarity, in this section we describe how we obtain the global translation assuming equally confident keypoints. We explain how we incorporate keypoint confidences in Section~\ref{sec:method:kpts}. 

To obtain the global translation $\mathbf{t}=[\tau_x, \tau_y,\tau_z]^T$ in a differentiable way, we derive a solution based on the Direct Linear Transform (DLT)~\cite{prince2012computer, hartley2003multiple}, adapted to our specific problem. Firstly, by design $\mathbf{K^{3D}}$ and $\mathbf{K^{2D}}$ give us a set of 2D-3D correspondences $\mathcal{M}=\{(k_i^{3D},k_i^{2D})\}_{i=1}^{N_K}$, with $k_i^{3D} = [x_i,y_i,z_i]^T$ and $k_i^{2D} = [u_i,v_i]^T$. Additionally, it is important to note that the 3D keypoints $\mathbf{K}^{3D}$ are expressed in a frame that shares the same orientation as the camera frame, with only the global root translation missing to map root-relative keypoint coordinates to camera-space coordinates. Assuming a pinhole camera model with intrinsic parameters $\mathbf{\Lambda}$, we can express the projection equation as:
\begin{align}
d_i \begin{bmatrix} u_i \\ v_i \\1 \end{bmatrix}
 &=
\mathbf{\Lambda}
\begin{bmatrix}
  1 & 0  & 0 & \aug & \tau_x \\
   0 & 1  & 0 & \aug & \tau_y \\
    0 & 0  & 1 & \aug & \tau_z \\
\end{bmatrix}
\begin{bmatrix} x_i \\ y_i \\z_i \\ 1  \end{bmatrix},
\end{align}
with $d_i$ the depth value of keypoint $i$.   Expanding and re-arranging, this gives a system of linear equations that can be written in the following form:
\begin{align}\label{sec:method:main_equation}
\begin{bmatrix}
  -1 & 0  & u_i'  \\
   0 & -1  & v_i'  \\
\end{bmatrix}
\begin{bmatrix} \tau_x \\\tau_y \\\tau_z \end{bmatrix}
 &=
\begin{bmatrix}
 x_i-z_iu_i' \\
    y_i-z_iv_i' \\
\end{bmatrix},
\end{align}
where
\begin{align}
\begin{bmatrix}
   u_i'  \\
   v_i'  \\
   1\\
\end{bmatrix}&=
\mathbf{\Lambda}^{-1}
\begin{bmatrix}
   u_i  \\
   v_i  \\
   1 \\ 
\end{bmatrix}.
\end{align}
Since Equation~\ref{sec:method:main_equation} is obtained considering a single keypoint correspondence, and since we have three unknowns, it is under-constrained. Using all the keypoints in our correspondence set, Equation~\ref{sec:method:main_equation} can be re-written as 
\begin{align}
\begin{bmatrix}
  -1 & 0  & u_1'  \\
    0 & -1  & v_1'  \\
  &\vdots &\\
   0 & -1  & v_{N_K}'  \\
\end{bmatrix}
\mathbf{t}
 &=
\begin{bmatrix}
 x_1-z_1u_1'\\
     y_1-z_1v_1'\\
    \vdots  \\
    y_{N_K}-z_{N_K}v_{N_K}'\\
\end{bmatrix},
\end{align}

which has the form $\mathbf{At}=\mathbf{B}$. To solve for $\mathbf{t}$, we consider the least-squares solutions 
    $\mathbf{t}^*=\argmin_\mathbf{t}||\mathbf{A}\mathbf{t}-\mathbf{B}||^2$~\cite{prince2012computer}, which can be obtained in closed-form:
\begin{equation}\label{sc:method:lsqsol}
\mathbf{t^*}= (\mathbf{A}^T\mathbf{A})^{-1}\mathbf{A}^T\mathbf{B}.
\end{equation}
The DGP module first uses the network outputs to build the matrices $\mathbf{A}$ and $\mathbf{B}$, and then uses the linear least-squares solution to solve for the translation. Since all the operations involved are differentiable, we can use this translation to obtain and backpropagate through camera-space vertex predictions, fully incorporating the root-finding task in an end-to-end training pipeline. 

\subsection{Keypoint Selection}
\label{sec:method:kpts}
While the solution presented in Section~\ref{sec:method:DLT} allows us to incorporate root finding into an end-to-end differentiable pipeline, it does not provide for any outlier filtering or keypoint selection mechanism that could help filter out more uncertain correspondences. This can be problematic in cases such as occlusion or self-occlusion of parts of the hand. In those instances, occluded parts would presumably result in more uncertain keypoint placements, they would be considered equally to visible keypoints when computing the global translation. To address this issue, we consider a weighted variant to our approach. Assuming each keypoint correspondence has a confidence score $w_i$ associated with it it -- in practice obtained from our weight decoder -- we first construct a weight matrix by duplicating each weight once and placing them in a diagonal matrix $\mathbf{W} = \diag([w_1,w_1, w_2,w_2 \hdots w_{N_K},w_{N_K}])$.
Then, we consider a weighted least-squares minimisation  $\mathbf{t}^*=\argmin_\mathbf{t}||\mathbf{W}(\mathbf{A}\mathbf{t}-\mathbf{B})||^2$, with closed-form solution:
\begin{equation}\label{sc:method:wlsqsol}
\mathbf{t^*}= (\mathbf{A}^T\mathbf{W}^2\mathbf{A})^{-1}\mathbf{A}^T\mathbf{W}^2\mathbf{B}.
\end{equation}

\subsection{Input Image Rectification} \label{sec:method:rectification}
Inspired by the recent work on monocular 3D geometry estimation from Yin \etal~\cite{yin2023metric3d}, the main idea is to establish a canonical camera space and transform all the training data to that space. During inference, the image is rectified just before entering the network, and the predictions are then mapped back to the original camera space. The original set of camera parameters is defined by $\{f, u_0, v_0\}$ where $f$ represents the focal length (we assume $f_x=f_y=f$) and $u_0$ and $v_0$ be the principal points. We resize the input image $\mathbf{I}$ with the ratio $\omega_r$ = $\frac{f^c}{f}$ which converts the camera parameters to $\{f^c, \omega_ru_0, \omega_rv_0\}$. Different to \cite{yin2023metric3d} and motivated by the hand being the object of interest, we further rectify the principal point to be the center of the hand crop, resulting in the final canonical intrinsics matrix $\mathbf{\Lambda}$ defined by $\{f^c, H/2, W/2\}$ where $f^c$ is the canonical focal length and the rectified principal point is the center of the input image. It is important to note that this does not affect the 3D geometry, thus root  $\mathbf{t}$ remains unchanged. 

\subsection{Architecture and Training Details}
\label{sec:method:nets}

\textbf{Network Overview.} In practice, our network first takes  an image $\mathbf{I}\in \mathbb{R}^{H \times W \times 3}$ as an input to  a convolutional encoder to produce a feature map $\mathbf{F}\in \mathbb{R}^{H/32\times W/32 \times C}$. The feature map $\mathbf{F}$ is then input to three separate decoder heads: a 2D decoder outputting a set of $N_K$ 2D keypoints $\mathbf{K}^{2D}$, a 3D decoder outputting the root-relative vertices $\mathbf{V}^{rel}$, and a weights decoder outputting a set of confidence weights $\mathbf{W}$.  Using $\mathbf{J}_{reg}$, we obtain the root-relative 3D keypoints $\mathbf{K}^{3D}$, forming a set of 2D-3D correspondences $\mathcal{M}=\{(u_i,v_i, x_i,y_i,z_i)\}_{i=1}^{N_K}$. Using $\mathcal{M}$, we then construct the matrices $\mathbf{A}$ and $\mathbf{B}$, and obtain $\mathbf{t}$ using Equations~\ref{sc:method:lsqsol}~and~\ref{sc:method:wlsqsol}. Finally, we use $\mathbf{t}$, $\mathbf{V}^{rel}$ and $\mathbf{K}^{3D}$ to obtain the camera-space vertices $\mathbf{V}^{cs}$ and camera-space keypoints,  following Equation~\ref{sec:method:eq_vcs}, and use all of the mentioned network outputs to construct our training losses.\\
\begin{wrapfigure}{r}{0.5\textwidth}
    \centering
    \includegraphics[width=0.5\textwidth]{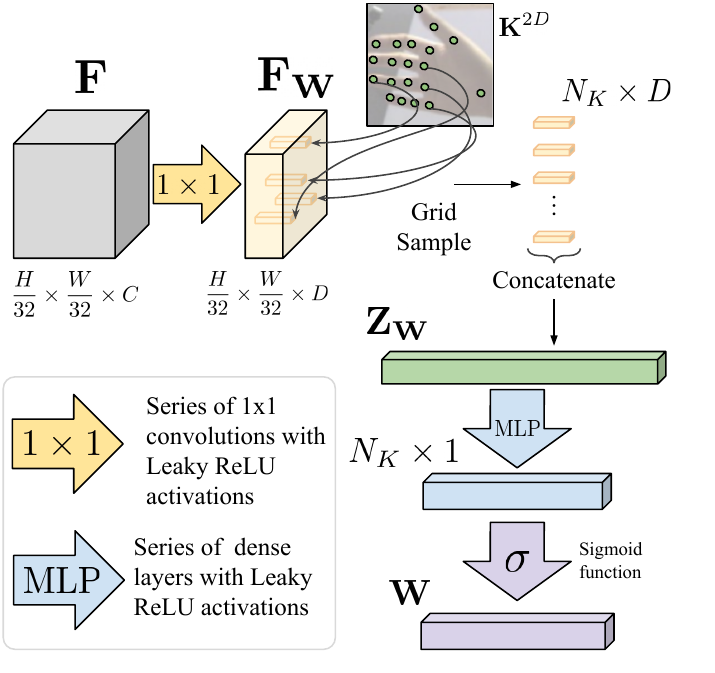}
    \caption{\textbf{Weight decoder head.}}
    \label{fig:weight_decoder}    
\end{wrapfigure}
\textbf{Weights decoder.} While our 2D and 3D decoders follow MobRecon~\cite{chen2022mobrecon}, our weights decoder is illustrated in Fig.~\ref{fig:weight_decoder}. Starting from the feature map $\mathbf{F} \in \mathbb{R}^{H/32\times W/32 \times C}$, we perform a series of $1 \times 1$ convolutions to obtain a new feature map $\mathbf{F_W} \in \mathbb{R}^{H/32\times W/32 \times D}$. We then use the 2D positions provided by $\mathbf{K}^{2D}$ in order to grid sample a set of $N_K$, $D$-dimentional features, which we concatenate in $D\times N_K$ dimensional latent vector $\mathbf{Z_w}$ which is then processed through a set of dense layers with leaky ReLU activations, and the final output is processed through a sigmoid function, forcing the  confidence weights to be in $[0,1]$. \\
\textbf{Losses.} We distinguish between two sets of losses: \textit{Relative-space} losses that are applied to any outputs that precede our global positioning module, and \textit{camera-space} losses that follow the global positioning module. \textit{Relative-space}: we follow MobRecon~\cite{chen2022mobrecon}, and use (1) a root-relative 3D Vertex $L_1$ Loss between the root-relative ground truth vertices and the outputs of the 3D decoder, (2) a 2D keypoint $L_1$ loss on the outputs of the 2D decoder, (3) a consistency loss that enforces consistency of  outputs between inputs with different visual augmentations, (4) a $L_1$ edge loss on the length of the predicted mesh edges, and (5) a normal loss on the predicted mesh normals.  For our \textit{camera-space} losses, we use (1) a root mean squared error loss on the translation, (2) a keypoint consistency loss, which is a $L_1$ loss between the outputs of our 2D decoder and the projection of the predicted camera-space 3D keypoints, and (3) a 2D vertex $L_1$ loss on the 2D projection of the 3D camera-space vertices. 

Implementation details and additional method information, including losses and network architecture details, are available in the supplementary material.

\section{Experiments}
\label{sec:experiments}

\textbf{Datasets.} We report our experiments on the following datasets: 
\begin{itemize}
    \item \textbf{FreiHAND}~\cite{zimmermann2019freihand}. We follow the evaluation protocol by \cite{chen2021camera, huang2023neural} using the  dataset for our experiments. The dataset consists of images, 3D hand poses and MANO~\cite{romero2022embodied} fittings. It provides 130,240 training and 3,960 test images. 
    \item \textbf{HO3D-v2}~\cite{hampali2020honnotate}. This dataset comprises real images capturing 3D hand-object interactions, with 66,034 images in the training set and 11,524 in the test set with MANO~\cite{romero2022embodied} model hand mesh annotations. Hands in this dataset suffer from severe occlusions caused by the manipulated object. The test set is not publicly available, and the evaluation is conducted on a public server. The server provides results in camera-space, root-relative, and aligned formats. However, participants are given ground truth camera-space hand translation values, and previous work typically reports results using this ground truth.
    \item \textbf{Human3.6M}~\cite{ionescu2013human3} This dataset is a large-scale 3D body pose benchmark containing 3.6 million frames with annotations of 3D joint coordinates and SMPL~\cite{pavlakos2019expressive} meshes. We follow existing evaluation protocols \cite{chen2021camera}, but do not use common aligned metrics and measure errors in camera-space. We adapt our framework to predict body meshes by just swapping MANO by SMPL.
\end{itemize}
\noindent\textbf{Metrics.} We report the following metrics:
\begin{itemize}
    \item \textbf{CS-MJE / CS-MVE}: Measures the error, in terms of Euclidean distance, between the predicted joints (MJE) / vertices (MVE) and the ground truth in camera-space (CS) coordinates. Both are average errors over the test set in mm. In some experiments, we compute the Area-Under-the-Curve (\textbf{AUC}) of Percentage of Correct Keypoints (\textbf{PCK}) vs. error thresholds.%, in our case PCK uses mesh vertices.
    \item \textbf{RS-MJE / RS-MVE}: This measures the error between Procrustes-aligned predicted and ground truth joints and vertices. This metric serves as a measure of root-relative reconstruction quality.
\end{itemize}
\subsection{Baselines and Method Ablations: Descriptions}
\label{sec:baselines}
\begin{table}[t]
\setlength{\tabcolsep}{3pt}
\resizebox{\textwidth}{!}{
\begin{tabular}{lcccccc}
\multirow{2}{*}[-2.em]{\textbf{Method}} &  \multicolumn{2}{c}{} &\multicolumn{2}{c}{\textbf{Root-Relative}} & \multicolumn{2}{c}{\textbf{Camera-Space}} \\\cmidrule(l){4-5} \cmidrule(l){6-7} 
& \begin{tabular}[c]{@{}c@{}}Image\\Rectification\end{tabular} & \begin{tabular}[c]{@{}c@{}}End-to-End \\Training \end{tabular} & {RS-MJE}$\downarrow$ & {RS-MVE}$\downarrow$ & {CS-MJE}$\downarrow$ & {CS-MVE}$\downarrow$ \\ \cmidrule(l){2-3} \cmidrule(r){1-1} \cmidrule(l){4-5} \cmidrule(l){6-7} 	
\textbf{B1.} Baseline + P\textit{n}P		&	\xmark	&	\xmark	&	6.8	&	6.9	&	50.1	&	50.2	\\
\textbf{B2.} Baseline + DLT		&	\xmark	&	\xmark	&	6.8	&	6.9	&	50.0	&	50.0	\\
\textbf{B3.} Baseline + RootNet~\cite{RootNet} 		&	\xmark	&	\xmark	&	6.8	&	6.9	&	62.6	&	62.5	\\
\textbf{B4.} Baseline + Optimization~\cite{chen2022mobrecon}		&	\xmark	&	\xmark	&	6.8	&	6.8	&	50.2	&	50.3	\\
\textbf{B5.} Root regression		&	\xmark	&	\cmark	&	7.2	&	7.6	&	81.3	&	81.8	\\\hdashline										
\rowcolor{lighterblue} \textbf{BR1.}  Baseline + DLT + Rect.		&	\cmark	&	\xmark	&	7.4	&	7.5	&	48.4	&	48.8	\\
\rowcolor{lighterblue}\textbf{BR2.}  Baseline + Optimization~\cite{chen2022mobrecon} + Rect. 		&	\cmark	&	\xmark	&	7.4	&	7.5	&	48.9	&	49.0	\\
\rowcolor{lighterblue} \textbf{BR3.}   Root regression + Rect.		&	\cmark	&	\cmark	&	7.8	&	7.7	&	52.6	&	55.4	\\\hdashline													
\textbf{A4.} Ours - w/o Rect. 		&	\xmark	&	\cmark	&	6.8	&	6.9	&	49.4	&	49.4	\\\hline									
\rowcolor{lightblue} Ours (Full framework)    		&	\cmark	&	\cmark	&	7.4	&	7.6	&	46.3	&	46.3	\\			\bottomrule														
\end{tabular}
}
\caption{\textbf{Baseline and ablation experiments on FreiHAND dataset}~\cite{zimmermann2019freihand}. The \textit{`Image Rectification' }column indicates whether the training images are rectified with our proposed approach. \textit{`End-to-end Training'} denotes whether gradients flow through the global positioning function during training. The task we care about for 3D interactions is quality in the camera space where we outperform all the baselines and variants.}
\label{table:baseline_freihand}
\end{table}

\textbf{Baselines.} Our root-relative module is combined with different methods for predicting camera-space root translation. Methods labeled as `Baseline + \{\textit{root prediction method}\}' incorporate the root-relative module (see Fig.~\ref{fig:main_fig}) without any global positioning system during training. Therefore, only the relative-space losses are applied and the positioning of the hand mesh in camera space occurs only at test time. In `\textbf{B1. Baseline + P\textit{n}P}', the global translation is obtained by solving the perspective-\textit{n}-point (P\textit{n}P) problem~\cite{prince2012computer}, using the outputs of our 2D decoder and the 3D root-relative keypoints as 2D-3D correspondences. `\textbf{B2. Baseline + DLT}' refers to our baseline wherein our formulation of the DLT module is applied at test time only during a forward pass, \textit{i.e.}, without gradient propagation through the network. In `\textbf{B3. Baseline + RootNet}', the global translation is determined using RootNet~\cite{RootNet}. `\textbf{B4. Baseline + Optimization}' corresponds to our reimplementation of MobRecon\cite{chen2022mobrecon}, where the global translation is identified via an optimization-based process that minimizes the re-projection errors of predicted keypoints. Finally, `\textbf{B5. Root Regression}' is the only end-to-end baseline that employs a decoder, similar to our weight decoder shown in Fig.\ref{fig:weight_decoder}, but for regressing the camera-space root value $\mathbf{t}$ and incorporating a translation loss. We also implemented some of these baselines with our image rectification step, denoted as \textbf{BR1-3}. Results are summarized in Table~\ref{table:baseline_freihand}.\\
\noindent\textbf{Ablations.} We conduct several ablation studies on our full pipeline, as shown in Table~\ref{table:ablation_others}. \textbf{Ours (Full Framework)} is our proposed method, which includes performing image rectification on the input image and conducting differentiable global positioning during training, with gradients back-propagating through the DGP module. `\textbf{A1. Ours - w/o DGP - w/o Rectification}' represents our framework without the proposed DGP and rectification steps, following the setup of B4. `\textbf{A2. Ours - w/o DGP}' evaluates the impact of our image rectification step in comparison to \textbf{A1}. In the `\textbf{A3. Ours - w/o Keypoint Weights $\mathbf{W}$}' experiment, we utilize our full framework but, instead of predicting keypoint confidences using our weight decoder, manually set all the weights to $1.0$. Lastly, in \textbf{A4. No Rectification + DGP} as detailed in Table~\ref{table:baseline_freihand}, we implement the full DGP module but exclude the image rectification from the pipeline.
\begin{table}[t]
\resizebox{\textwidth}{!}{
\begin{tabular}{@{}lcccccc@{}}
\multirow{2}{*}[-1.em]{\textbf{Method}} &  \multicolumn{2}{c}{\textbf{FreiHAND}} &\multicolumn{2}{c}{\textbf{HO3D-v2}} & \multicolumn{2}{c}{\textbf{Human 3.6M}} \\ \cmidrule(l){2-3} \cmidrule(l){4-5} \cmidrule(l){6-7} 
& CS-MJE$\downarrow$ & CS-MVE$\downarrow$ & CS-MJE$\downarrow$ & CS-MVE$\downarrow$  & CS-MJE$\downarrow$  & CS-MVE$\downarrow$\\\cmidrule(l){2-3} \cmidrule(r){1-1} \cmidrule(l){4-5} \cmidrule(l){6-7} 										
\textbf{A1.} Ours -  w/o DGP - w/o rectification		&	50.2	&	50.3	&	121.7	&	121.6	&	336.9	&	342.8	\\
\textbf{A2.} Ours - w/o DGP		&	48.9	&	49.0	&	85.3	&	85.4	&	164.7	&	179.3	\\
\textbf{A3.} Ours - w/o keypoint weights $\mathbf{W}$		&	46.6	&	46.6	&	50.4	&	50.4	&	159.9	&	174.0	\\
\cmidrule(l){2-3} \cmidrule(r){1-1} \cmidrule(l){4-5} \cmidrule(l){6-7}					
\rowcolor{lightblue}\textbf{Ours} (Full framework)    		&	46.3	&	46.3	&	50.3	&	50.3	&	147.6	&	162.0	\\
\bottomrule
\end{tabular}
}
\caption{\textbf{Method ablation experiments}. We outperform all ablations on camera-space predictions across all datasets, validating our design choices and use of HandDGP.}
\label{table:ablation_others}
\end{table}
\subsection{Baselines and Method Ablations: Results Discussion}
\noindent\textbf{Learning the 3D Global Positioning Function.} In this evaluation, we assess the effect of back-propagating gradients from the DGP to the network, as proposed in Section~\ref{sec:method:DLT}. 
In Table~\ref{table:baseline_freihand} we observe that DGP outperforms all the other root-finding mechanisms. Non-learning approaches such as P\textit{n}P, DLT and the optimization from Chen~\etal\cite{chen2022mobrecon} (\textbf{B1, \textbf{B2}, B4}) offer better predictions than learning-based methods such as RootNet~\cite{RootNet} and our regression baseline (\textbf{B5}). These results still hold when image rectification is applied (\textbf{BR1-3}). Notably, the DGP formulation, even without training (forward pass), achieves comparable results to more sophisticated optimization methods, validating our design choice. In Table~\ref{table:ablation_others} we observe a significant reduction in error in all the datasets when learning the global positioning function with the proposed  DGP (\textbf{A2}). For example, the camera-space vertex error is reduced by $2.7$ mm, $35.1$ mm and $17.3$ mm in FreiHAND, HO3D-v2 and Human3.6M datasets respectively. This last result shows that HandDGP also transfers well to full body mesh predictions.\\
 \begin{wrapfigure}{r}{0.4\textwidth}
    \centering
    \includegraphics[width=0.4\textwidth]{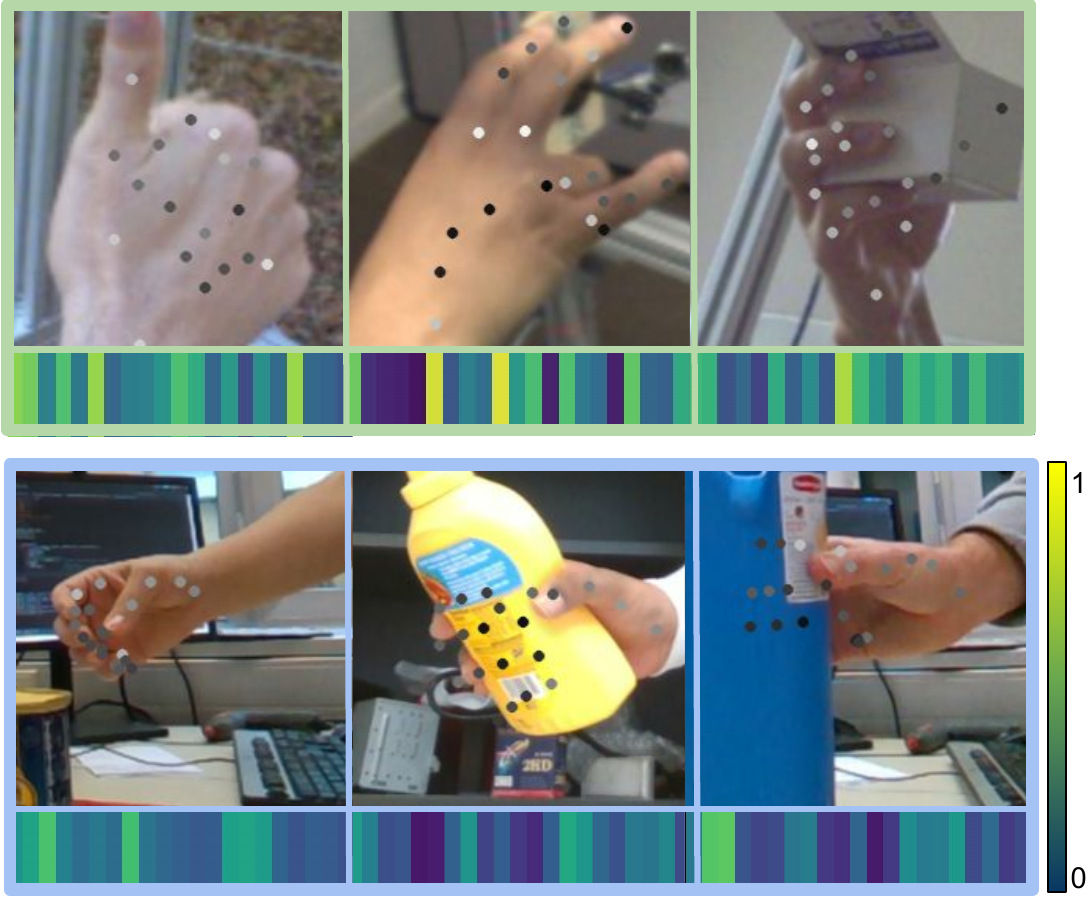}
    \caption{\textbf{Keypoint selection.} Effect of keypoint selection with our weight decoder. Test-set images on {\color{ForestGreen}FreiHAND} and  {\color{Cerulean}HO3D-v2} with the 2D keypoints overlaid: The brighter the keypoint, the higher the weight.}
    \label{fig:keypoint_illustration}
\end{wrapfigure}
\noindent\textbf{Image Rectification Effect.} We observe that rectifying input images enhances the learning of scale-sensitive features, as we remove one source of ambiguity by keeping the camera intrinsics constant during training. Interestingly, we note that while rectifying images aids in camera space predictions, it impacts root-relative predictions (\textbf{Ours} vs \textbf{A4}). We hypothesize that the primary reason for this degradation is that rectifying training images \textit{de facto} reduces the amount of data augmentations that the network encounters during training, affecting the prediction of both 2D keypoints and 3D mesh geometry. This suggests a potential trade-off between root-relative and absolute prediction quality, depending on how the training data is processed, exacerbated by the 2D-to-3D depth and scale ambiguity. The network may generate incorrect hand shapes to compensate for errors in translation prediction and vice versa. 
In all cases, better camera-space positioning is achieved when the input image is rectified.\\

\noindent\textbf{Keypoint Selection.} In Table~\ref{table:ablation_others} we observe that learning the keypoint selection weights $\mathbf{W}$ introduced in Eq.~\ref{sc:method:wlsqsol}, instead of leaving them fixed  further helps reducing the prediction error (\textbf{Ours} vs. \textbf{A3}). We illustrate this effect qualitatively in Fig.~\ref{fig:keypoint_illustration}. For images in the test set, we draw the keypoints predicted by our 2D decoder onto the input image, color-coding each keypoint using its associated weight, as output by our weight decoder. The brighter the keypoint, the higher the weight associated with it. We also illustrate the weights themselves associated with each image in the bottom row of the figure. The first observation is that the weights associated with each keypoint clearly vary from image to image, indicating that our weight decoder actively contributes to our overall pipeline. Next, we notice that occluded keypoints generally tend to be associated with lower weights, suggesting that our method tends to focus on higher confidence keypoints.\\

\begin{figure}[t]
    \centering
    \includegraphics[width=\textwidth, trim={0.0cm, 24cm, 3.7cm, 0.0cm}, clip]{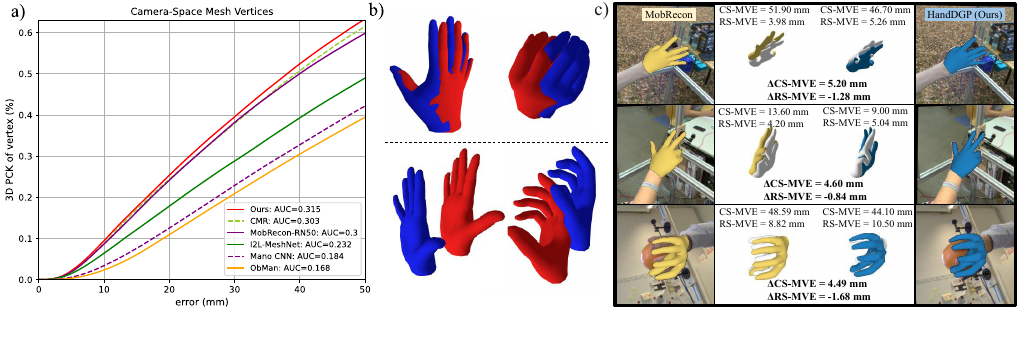}
    \caption{\textbf{(a)} \textbf{3D PCK} for camera-space hand mesh prediction on FreiHAND. \textbf{(b) Camera-space hand mesh predictions} rotated for illustration purposes. All meshes project correctly in the image, however some predictions display a 3D error offset. \textbf{(c) Root-relative vs camera-space errors}. Selected FreiHAND images with average camera-space (CS) and root-relative (RS) errors and ground truth (mesh in white).}
    \label{fig:curve-pck}
\end{figure}

\begin{figure}[t]
    \centering
    \includegraphics[width=1\textwidth]{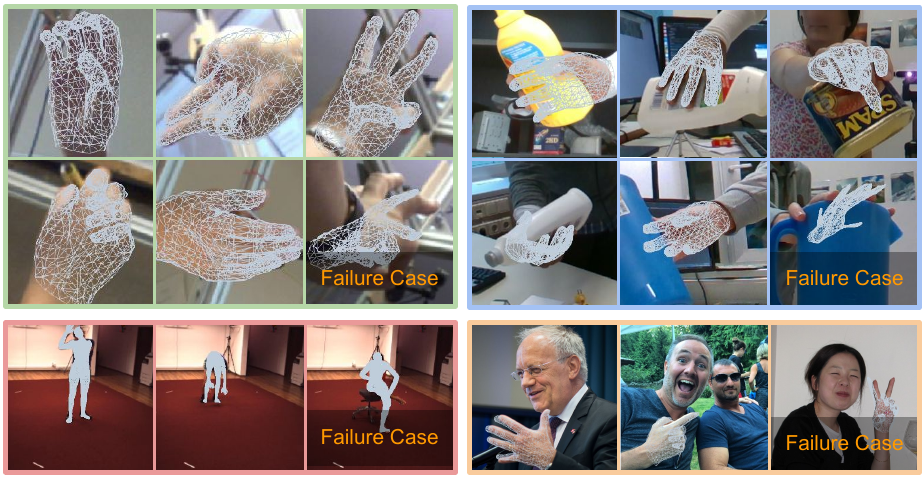}
    \caption{\textbf{Qualitative visualizations} of our camera-space mesh prediction framework on {\color{ForestGreen}FreiHAND}, {\color{Cerulean}HO3D-v2}, {\color{OrangeRed}Human3.6M} and {\color{Goldenrod}in-the-wild} web images~\cite{kuznetsova2020open}. Meshes are projected into the image plane using perspective projection, including failure cases. Further qualitative results and comparisons are available in the supplementary material.}
    \label{fig:qualitative}
\end{figure}

\noindent\textbf{Qualitative Results.} In  Fig.~\ref{fig:qualitative} we present qualitative results using our full framework on test images from different datasets. In most cases, the hand projections appear accurate, as also quantitatively demonstrated in our relative results in Table~\ref{table:baseline_freihand}. Additionally, we showcase some rare failure cases where high ambiguity arises due to blurring, changes in viewpoint, and self-occlusion. Merely displaying projected 2D meshes might not convey the complete truth, as predictions are made in the actual 3D camera space. To address this, we include predicted 3D meshes from rotated viewpoints to illustrate the actual translation gap between the predictions and the ground truth in Fig.~\ref{fig:curve-pck} (b). While the hand posture is generally correct, we observe instances of translation errors, sometimes overcompensating with larger or smaller hand shapes.\\ 

\noindent\textbf{Root-relative results discussion.} While not the focus of this work, Table~\ref{table:baseline_freihand} shows our method is outperformed by some baselines in the root-relative task by an average of 0.6 mm (RS-MVE). However, our improvements on camera-space errors (CS-MVE) are about 4 mm. To assess if this RS-MVE increase is an acceptable trade-off, Fig.~\ref{fig:curve-pck} (c) compares our baseline and HandDGP on the FreiHAND test set. We selected images with a CS-MVE decrease of about 4 mm and an RS-MVE increase of at least 0.6 mm for MobRecon over HandDGP. The CS-MVE improvement is more evident when rotating the 3D viewpoint (columns 2-3). Our meshes are visibly closer to the ground truth, crucial for applications requiring interaction with real and digital objects. Higher RS-MVE (0.84-1.68 mm) are harder to visualize. Notably, a better RS error doesn't always result in a better image projection (last row) and can hide scale errors due to Procrustes alignment. We believe the 4 mm CS-MVE improvement is significant, and our visualizations suggest a 0.6 mm increase in RS errors is an acceptable trade-off.\\

\noindent\textbf{Framework Generalizability.} In the supp. material we show the benefits of applying our framework to a different state-of-the-art root-relative  method~\cite{chen2021camera}.\\
\subsection{Comparison with State of the Art}
\label{sec:sota}
\begin{wraptable}{r}{0.4\textwidth}
\resizebox{0.4\textwidth}{!}{
\centering
\captionsetup{width=0.9\textwidth}
\begin{tabular}{ccc}
\hline
\multirow{2}{*}{\vspace{2.5ex}Method}   & {CS-MJE}$\downarrow$ & {CS-MVE}$\downarrow$ \\ 
\hline
ObMan~\cite{hasson19_obman}                                 & 85.2 &    85.2        \\
MANO CNN~\cite{zimmermann2019freihand}                     & 71.3 &    71.5        \\
I2L-MeshNet~\cite{I2L}                            & 60.3  &    60.4      \\
NVF~\cite{huang2023neural}                             & \underline{47.2} &  n/a$^\dagger$ \\
\hline
\rowcolor{lightergray} 
CMR-SG-RN18~\cite{chen2021camera}                          & 49.7  &    49.8     \\
\rowcolor{lightergray} 
CMR-SG-RN50~\cite{chen2021camera}                         & 48.8  &    \underline{48.9}      \\
\rowcolor{lightergray} 
MobRecon-RN50~\cite{chen2022mobrecon}                                           & 50.2 &  50.3 \\       
\hline
\rowcolor{lightblue} 
Ours                                      & \textbf{46.3} & \textbf{46.3}  \\
\hline
\end{tabular}}
\caption{\textbf{State of the art comparison on FreiHAND}. $^\dagger$can only be evaluated for keypoints. }
\label{table:sota_freihand}
\end{wraptable}
\textbf{FreiHAND:} In Table~\ref{table:sota_freihand} and Fig.~\ref{fig:curve-pck} (a) we compare our proposed framework with state-of-the-art camera-space methods. Notably, our method achieves the lowest camera-space errors among all methods. We achieve a 2.6 mm error improvement over  CMR~\cite{chen2021camera}, despite also using segmentation masks in their root finding. 
We also compare with the ResNet50 variant of MobRecon~\cite{chen2022mobrecon} which is the same as our baseline B4 from the previous section. 
For MobRecon, we report our results based on our implementation, which closely aligns with the one provided by the authors, with slight changes in the data processing that are also shared with our method. A similar trend is observed in Figure~\ref{fig:curve-pck} where our method achieves the highest AUC of vertex PCK's among all methods closely followed by both CMR and MobRecon-RN-50. We also compare favorably with NVF~\cite{huang2023neural}, the only available method that, similar to us, predicts directly in camera space. Note that NVF does not directly predict hand meshes, making them not easily comparable due to the loss of the MANO topology as their meshes are generated using Marching Cubes. We observe that the root-relative + 2D-3D global positioning paradigm (ourselves, CMR and MobRecon) performs significantly better than other methods, followed by I2L-MeshNet~\cite{I2L} that uses~\cite{RootNet} for root positioning, similar to our baseline B3.

\begin{wraptable}{l}{0.4\textwidth}	
\resizebox{0.4\textwidth}{!}{
\centering					
\captionsetup{width=0.9\textwidth}					
\begin{tabular}{ccc}					
\hline					
\multirow{2}{*}{\vspace{2.5ex}Method}   & {CS-MJE}$\downarrow$ & {CS-MVE}$\downarrow$ \\					
\hline		
HandOccNet~\cite{park2022handoccnet}	&	156.4	&	156.2	\\
MobRecon-RN50~\cite{chen2022mobrecon}	&	121.7	&	121.6	\\
Hasson \etal~\cite{hasson2020leveraging}	&	\underline{55.2}	&	\underline{55.1}	\\
\hline					
\rowcolor{lightblue} 					
Ours	&	\textbf{50.3}	&	\textbf{50.3}	\\
\hline					
\end{tabular}
}
\caption{\textbf{State of the art comparison on HO3D-v2.}}			
\label{table:ho3d}					
\end{wraptable}		
\textbf{HO3D-v2:} In Table~\ref{table:ho3d} we present quantitative results of our method compared to state-of-the-art methods in camera-space coordinates using the public submission server. It is important to note that previous work typically reports their results in relative coordinates after an aligning step and often uses provided ground-truth root values. We do not have a way to know which participants on the leaderboard used this ground-truth; because of this, we had to recompute the scores. We report results of the available subset of methods that: i) have publicly available code, ii) provide a trained model, \textbf{and} iii) include a global coordinate prediction stage. This dataset is more challenging than FreiHAND as it is object-focused, and several occlusions are present. We observe that our method compares favorably to HandOccNet\cite{park2022handoccnet}, which is currently the state of the art in relative pose predictions. To compute these predictions, we run their code that performs test-time optimization to predict the root translation. Despite their mesh projections looking good, the 3D predictions are often incorrect, with errors in the order of 15 cm. We also compare favorably to the work of Hasson~\etal~\cite{hasson2020leveraging} that predicts both object and hand global translations. Given that object size is constant—compared to hands—predicting object roots is likely to help with the scale/depth ambiguity problem. We show qualitative comparisons both in the supplementary material and in the video presentation.

\section{Conclusion}
\label{sec:conclusion}
We presented a framework for camera space hand mesh prediction, enabling learning directly in camera space. Our baseline and ablation studies validated our design choices, showing our method surpasses state-of-the-art approaches that predict hand meshes in camera-space coordinates. Estimating absolute 3D geometry from a single RGB image is inherently ill-posed. Rectifying images and predicting in camera space help reduce errors.

Our experiments show that while root-relative error is in the low single-digit millimeters, likely lower than annotation error, camera space error is 6 to 7 times larger. This is visually illustrated in Fig.~\ref{fig:curve-pck} (b) and (c), suggesting a significant portion of total errors stems from 2D-to-3D depth ambiguity. We conjecture that isolating the hand from its context will soon reach a performance ceiling. Further research in new datasets and context-aware approaches, such as using scene geometry or objects, is needed to advance camera-space mesh inference.
\clearpage
\section*{Acknowledgements}
We would like to thank Filippo Aleotti for his help with baseline experiments and infrastructure;  Jamie Watson, Zawar Qureshi, and Jakub Powierza for their help with infrastructure; Axel Laguna for his insightful discussions on minimal solvers and network architectures; Daniyar Turmukhambetov for valuable technical discussions; and Gabriel Brostow, Sara Vicente, Jessica Van Brummelen, and Michael Firman for their valuable feedback on different versions of the manuscript.

{
    \small
    \bibliographystyle{splncs04}
    \bibliography{main}

\begin{thebibliography}{10}
\providecommand{\url}[1]{\texttt{#1}}
\providecommand{\urlprefix}{URL }
\providecommand{\doi}[1]{https://doi.org/#1}

\bibitem{antotsiou2018task}
Antotsiou, D., Garcia-Hernando, G., Kim, T.K.: Task-oriented hand motion retargeting for dexterous manipulation imitation. In: ECCV Workshop (2018)

\bibitem{AVP}
Apple: {Vision Pro}. \url{https://www.apple.com/apple-vision-pro/}, [Online; accessed 7-March-2024]

\bibitem{armagan2020measuring}
Armagan, A., Garcia-Hernando, G., Baek, S., Hampali, S., Rad, M., Zhang, Z., Xie, S., Chen, M., Zhang, B., Xiong, F., et~al.: Measuring generalisation to unseen viewpoints, articulations, shapes and objects for 3{D} hand pose estimation under hand-object interaction. In: ECCV (2020)

\bibitem{baek2019pushing}
Baek, S., Kim, K.I., Kim, T.K.: Pushing the envelope for rgb-based dense 3d hand pose estimation via neural rendering. In: CVPR (2019)

\bibitem{baek2020weakly}
Baek, S., Kim, K.I., Kim, T.K.: Weakly-supervised domain adaptation via gan and mesh model for estimating 3{D} hand poses interacting objects. In: CVPR (2020)

\bibitem{bhatnagar2020loopreg}
Bhatnagar, B.L., Sminchisescu, C., Theobalt, C., Pons-Moll, G.: Loopreg: Self-supervised learning of implicit surface correspondences, pose and shape for 3d human mesh registration. In: NeurIPS (2020)

\bibitem{bhowmik2020reinforced}
Bhowmik, A., Gumhold, S., Rother, C., Brachmann, E.: Reinforced feature points: Optimizing feature detection and description for a high-level task. In: CVPR (2020)

\bibitem{boukhayma20193d}
Boukhayma, A., Bem, R.d., Torr, P.H.: 3d hand shape and pose from images in the wild. In: CVPR (2019)

\bibitem{brachmann2017dsac}
Brachmann, E., Krull, A., Nowozin, S., Shotton, J., Michel, F., Gumhold, S., Rother, C.: Dsac-differentiable ransac for camera localization. In: CVPR (2017)

\bibitem{cao2021reconstructing}
Cao, Z., Radosavovic, I., Kanazawa, A., Malik, J.: Reconstructing hand-object interactions in the wild. In: CVPR (2021)

\bibitem{chen2020end}
Chen, B., Parra, A., Cao, J., Li, N., Chin, T.J.: End-to-end learnable geometric vision by backpropagating pnp optimization. In: CVPR (2020)

\bibitem{epropnp}
Chen, H., Wang, P., Wang, F., Tian, W., Xiong, L., Li, H.: Epro-pnp: Generalized end-to-end probabilistic perspective-n-points for monocular object pose estimation. In: CVPR (2022)

\bibitem{chen2021i2uv}
Chen, P., Chen, Y., Yang, D., Wu, F., Li, Q., Xia, Q., Tan, Y.: I2uv-handnet: Image-to-uv prediction network for accurate and high-fidelity 3d hand mesh modeling. In: ICCV (2021)

\bibitem{chen2022mobrecon}
Chen, X., Liu, Y., Dong, Y., Zhang, X., Ma, C., Xiong, Y., Zhang, Y., Guo, X.: Mobrecon: Mobile-friendly hand mesh reconstruction from monocular image. In: CVPR (2022)

\bibitem{chen2021camera}
Chen, X., Liu, Y., Ma, C., Chang, J., Wang, H., Chen, T., Guo, X., Wan, P., Zheng, W.: Camera-space hand mesh recovery via semantic aggregation and adaptive 2{D}-1{D} registration. In: CVPR (2021)

\bibitem{chen2023hand}
Chen, X., Wang, B., Shum, H.Y.: Hand avatar: Free-pose hand animation and rendering from monocular video. In: CVPR (2023)

\bibitem{chen2021model}
Chen, Y., Tu, Z., Kang, D., Bao, L., Zhang, Y., Zhe, X., Chen, R., Yuan, J.: Model-based 3d hand reconstruction via self-supervised learning. In: CVPR (2021)

\bibitem{garcia2020physics}
Garcia-Hernando, G., Johns, E., Kim, T.K.: Physics-based dexterous manipulations with estimated hand poses and residual reinforcement learning. In: IROS (2020)

\bibitem{ge20193d}
Ge, L., Ren, Z., Li, Y., Xue, Z., Wang, Y., Cai, J., Yuan, J.: 3d hand shape and pose estimation from a single rgb image. In: CVPR (2019)

\bibitem{hampali2020honnotate}
Hampali, S., Rad, M., Oberweger, M., Lepetit, V.: Honnotate: A method for 3{D} annotation of hand and object poses. In: CVPR (2020)

\bibitem{hampali2022keypoint}
Hampali, S., Sarkar, S.D., Rad, M., Lepetit, V.: Keypoint transformer: Solving joint identification in challenging hands and object interactions for accurate 3{D} pose estimation. In: CVPR (2022)

\bibitem{han2020megatrack}
Han, S., Liu, B., Cabezas, R., Twigg, C.D., Zhang, P., Petkau, J., Yu, T.H., Tai, C.J., Akbay, M., Wang, Z., et~al.: Megatrack: monochrome egocentric articulated hand-tracking for virtual reality. ACM TOG  (2020)

\bibitem{hartley2003multiple}
Hartley, R., Zisserman, A.: Multiple view geometry in computer vision. Cambridge university press (2003)

\bibitem{hasson2020leveraging}
Hasson, Y., Tekin, B., Bogo, F., Laptev, I., Pollefeys, M., Schmid, C.: Leveraging photometric consistency over time for sparsely supervised hand-object reconstruction. In: CVPR (2020)

\bibitem{hasson19_obman}
Hasson, Y., Varol, G., Tzionas, D., Kalevatykh, I., Black, M.J., Laptev, I., Schmid, C.: Learning joint reconstruction of hands and manipulated objects. In: CVPR (2019)

\bibitem{huang2023neural}
Huang, L., Lin, C.C., Lin, K., Liang, L., Wang, L., Yuan, J., Liu, Z.: Neural voting field for camera-space 3{D} hand pose estimation. In: CVPR (2023)

\bibitem{ionescu2013human3}
Ionescu, C., Papava, D., Olaru, V., Sminchisescu, C.: Human3.6{M}: Large scale datasets and predictive methods for 3{D} human sensing in natural environments. TPAMI  (2013)

\bibitem{iqbal2018hand}
Iqbal, U., Molchanov, P., Breuel Juergen~Gall, T., Kautz, J.: Hand pose estimation via latent 2.5{D} heatmap regression. In: ECCV (2018)

\bibitem{kanazawa2018end}
Kanazawa, A., Black, M.J., Jacobs, D.W., Malik, J.: End-to-end recovery of human shape and pose. In: CVPR (2018)

\bibitem{karunratanakul2021skeleton}
Karunratanakul, K., Spurr, A., Fan, Z., Hilliges, O., Tang, S.: A skeleton-driven neural occupancy representation for articulated hands. In: 3DV (2021)

\bibitem{karunratanakul2020grasping}
Karunratanakul, K., Yang, J., Zhang, Y., Black, M.J., Muandet, K., Tang, S.: Grasping field: Learning implicit representations for human grasps. In: 3DV (2020)

\bibitem{kulon2020weakly}
Kulon, D., Guler, R.A., Kokkinos, I., Bronstein, M.M., Zafeiriou, S.: Weakly-supervised mesh-convolutional hand reconstruction in the wild. In: CVPR (2020)

\bibitem{kuznetsova2020open}
Kuznetsova, A., Rom, H., Alldrin, N., Uijlings, J., Krasin, I., Pont-Tuset, J., Kamali, S., Popov, S., Malloci, M., Kolesnikov, A., et~al.: The open images dataset v4: Unified image classification, object detection, and visual relationship detection at scale. IJCV  (2020)

\bibitem{li2022cliff}
Li, Z., Liu, J., Zhang, Z., Xu, S., Yan, Y.: Cliff: Carrying location information in full frames into human pose and shape estimation. In: ECCV (2022)

\bibitem{lin2021end}
Lin, K., Wang, L., Liu, Z.: End-to-end human pose and mesh reconstruction with transformers. In: CVPR (2021)

\bibitem{lin2021mesh}
Lin, K., Wang, L., Liu, Z.: Mesh graphormer. In: ICCV (2021)

\bibitem{quest3}
Meta: {Quest 3}. \url{https://www.meta.com/us/quest/quest-3/}, [Online; accessed 7-March-2024]

\bibitem{mihajlovic2021leap}
Mihajlovic, M., Zhang, Y., Black, M.J., Tang, S.: Leap: Learning articulated occupancy of people. In: CVPR (2021)

\bibitem{RootNet}
Moon, G., Chang, J.Y., Lee, K.M.: Camera distance-aware top-down approach for 3{D} multi-person pose estimation from a single {RGB} image. In: ICCV (2019)

\bibitem{I2L}
Moon, G., Lee, K.M.: {I2L-MeshNet}: Image-to-lixel prediction network for accurate 3{D} human pose and mesh estimation from a single {RGB} image. In: ECCV (2020)

\bibitem{park2022handoccnet}
Park, J., Oh, Y., Moon, G., Choi, H., Lee, K.M.: Handoccnet: Occlusion-robust 3d hand mesh estimation network. In: CVPR (2022)

\bibitem{pavlakos2019expressive}
Pavlakos, G., Choutas, V., Ghorbani, N., Bolkart, T., Osman, A.A., Tzionas, D., Black, M.J.: Expressive body capture: 3{D} hands, face, and body from a single image. In: CVPR (2019)

\bibitem{peng2021neural}
Peng, S., Zhang, Y., Xu, Y., Wang, Q., Shuai, Q., Bao, H., Zhou, X.: Neural body: Implicit neural representations with structured latent codes for novel view synthesis of dynamic humans. In: CVPR (2021)

\bibitem{prince2012computer}
Prince, S.J.: Computer vision: models, learning, and inference. Cambridge University Press (2012)

\bibitem{remelli2020lightweight}
Remelli, E., Han, S., Honari, S., Fua, P., Wang, R.: Lightweight multi-view 3{D} pose estimation through camera-disentangled representation. In: CVPR (2020)

\bibitem{romero2022embodied}
Romero, J., Tzionas, D., Black, M.J.: Embodied hands: Modeling and capturing hands and bodies together. ACM TOG  (2017)

\bibitem{saito2019pifu}
Saito, S., Huang, Z., Natsume, R., Morishima, S., Kanazawa, A., Li, H.: Pifu: Pixel-aligned implicit function for high-resolution clothed human digitization. In: CVPR (2019)

\bibitem{spurr2020weakly}
Spurr, A., Iqbal, U., Molchanov, P., Hilliges, O., Kautz, J.: Weakly supervised 3d hand pose estimation via biomechanical constraints. In: ECCV (2020)

\bibitem{tang2021towards}
Tang, X., Wang, T., Fu, C.W.: Towards accurate alignment in real-time 3{D} hand-mesh reconstruction. In: ICCV (2021)

\bibitem{wei2023generalized}
Wei, T., Patel, Y., Shekhovtsov, A., Matas, J., Barath, D.: Generalized differentiable ransac. In: ICCV (2023)

\bibitem{yin2023metric3d}
Yin, W., Zhang, C., Chen, H., Cai, Z., Yu, G., Wang, K., Chen, X., Shen, C.: Metric{3D}: Towards zero-shot metric 3{D} prediction from a single image. In: ICCV (2023)

\bibitem{yuan2018depth}
Yuan, S., Garcia-Hernando, G., Stenger, B., Moon, G., Chang, J.Y., Lee, K.M., Molchanov, P., Kautz, J., Honari, S., Ge, L., et~al.: Depth-based 3{D} hand pose estimation: From current achievements to future goals. In: CVPR (2018)

\bibitem{zhang2021hand}
Zhang, X., Huang, H., Tan, J., Xu, H., Yang, C., Peng, G., Wang, L., Liu, J.: Hand image understanding via deep multi-task learning. In: ICCV (2021)

\bibitem{zhang2019end}
Zhang, X., Li, Q., Mo, H., Zhang, W., Zheng, W.: End-to-end hand mesh recovery from a monocular {RGB} image. In: ICCV (2019)

\bibitem{zhou2020monocular}
Zhou, Y., Habermann, M., Xu, W., Habibie, I., Theobalt, C., Xu, F.: Monocular real-time hand shape and motion capture using multi-modal data. In: CVPR (2020)

\bibitem{zimmermann2019freihand}
Zimmermann, C., Ceylan, D., Yang, J., Russell, B., Argus, M., Brox, T.: Frei{HAND}: A dataset for markerless capture of hand pose and shape from single {RGB} images. In: ICCV (2019)

\end{thebibliography}
}

\end{document}